# A Generative Adversarial Approach To ECG Synthesis And Denoising

Karol Antczak
*Institute of Computer and Information Systems*
*Military University Of Technology*
Warsaw, Poland
karol.antczak@wat.edu.pl

*Abstract*— Generative Adversarial Networks (GAN) are known to produce synthetic data that are difficult to discern from real ones by humans. In this paper we present an approach to use GAN to produce realistically looking ECG signals. We utilize them to train and evaluate a denoising autoencoder that achieves state-of-the-art filtering quality for ECG signals. It is demonstrated that generated data improves the model performance compared to the model trained on real data only. We also investigate an effect of transfer learning by reusing trained discriminator network for denoising model.

## I. INTRODUCTION

Measurement of electrical activity of the heart plays a basic role in modern diagnostics. Of multiple electrophysiological parameters of the heart, one of the most informative is the change in voltage over time caused by depolarization and repolarization of the heart muscle. A process that produces a graphical representation of it is known as electrocardiography, while the representation itself – an electrocardiogram (ECG). There are several characteristic entities that can be identified on ECG: QRS complex and P, T and U waves – each associated with the specific phenomena occurring during the single cardiac cycle (see Figure 1). Analysis of these entities associated with the knowledge of electrocardiogram scale allows, among other things, to calculate heart rate and to detect rhythm abnormalities such as atrial fibrillation, atrial flutter, cardiac arrhythmia, sinus tachycardia and sinus bradycardia. Shape analysis allows to detect another group of diseases; for example, axis deviation of QRS complex is a symptom of ventricular hypertrophy, anterior and posterior fascicular block and others.

In order to maximize the diagnostic value of ECG, it should contain only diagnostically relevant components. However, the real ECG contains a noise caused by multiple factors [1]. The most common are baseline wander, motion artifacts, electromyography (EMG) noise and power line interference. Furthermore, characteristic frequencies of noises often overlap with the frequency of the useful components of ECG. This makes filtering the signal a challenging task. A number of methods were developed over years, designed mostly for filtering specific kinds of noise [1], [2]. Examples include bandpass/notch filters (used primarily to remove power line interference), wavelet transform-based methods (applied for motion artifacts and EMG removal) and empirical mode decomposition (preferred for filtering out baseline wandering noise).

In the twenty-first century, apart from classical filtering methods, techniques based on artificial neural networks also appeared. One of the first applications was presented by Moein [3]. It was a multi-layer perceptron trained to imitate Kalman low-pass filter. Since then, the development of deep learning and greater availability of training data (in a form of medical databases) have allowed to use more advanced models. The temporal nature of ECG signal makes it a natural candidate for processing using recurrent neural networks. A research from 2018 have investigated Long Short-Term Memory (LSTM)-based network for the purpose of denoising, outperforming classical filters in case of heavily noised signal [4]. Single-layer LSTM networks were also explored for ECG classification [5]. Nonetheless, it seems that recurrent models have recently fallen out of favor in the area of sequence processing, being outperformed by convolution-based networks [6], [7]. More recently, a convolutional neural network have been presented for denoising of fetal ECG [8], reporting an increase in correlation coefficient between original and corrupted signal from 0.6 to 0.8 .

## II. PROPOSED APPROACH

We introduce a deep model for ECG denoising, trained with the help of artificial signals generated by GAN. To achieve this, we build several auxiliary neural models. First, we use real ECG data to train 2D convolutional network that operates on time-frequency representations of signals to classify them according to the diagnostic group. This allows us to compute an equivalent of Inception score which expresses a level of data "realism" that correlates well with human perception. Next, we train Wasserstein GAN operating in time-domain to produce artificial data that achieve possibly high Inception score. Generated artificial data are then processed using parametric model of noise. Finally, we train a denoising convolutional autoencoder that reconstructs initial, noise-free signal based on noisy input in time-domain. We evaluate it using both artificial and real data and various performance metrics. In summary, our approach for the development of the denoising model includes following steps:

1. Train the Inception network for evaluation of ECG signal quality.
2. Train the GAN for generating artificial ECG signals, using Inception score for selecting the best model.
3. Generate signal dataset with GAN and process it using the parametric noise model.
4. Train the denoising autoencoder using pairs of noised and clear signals.

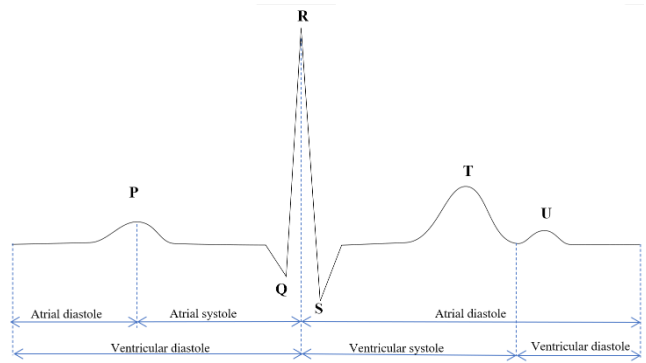

Figure 1. Anatomy of ECG cycle.

### III. RELATED RESEARCH

Generative Adversarial Networks (GAN) are deep learning models that produce artificial data imitating their real-life counterparts [9]. They comprise of a pair of networks, called generator and discriminator, that learn by playing a competitive minimax game. The generator produces data that is evaluated by the discriminator, along with samples of real data. The discriminator tries to determine which samples are generated and which are real, while the generator tries to fool it by producing more and more realistic samples. It has been shown that such game has (theoretically) a non-dominated solution which minimizes Jensen-Shannon divergence between generated and real data [9]. However, "vanilla" GANs have several issues, the most vital being the mode collapse (i.e. reduction of the output data diversity) and long convergence time. Both of these problems were partially resolved by models such as Wasserstein GAN [10] that minimizes smoother Wasserstein-1 distance rather than J-S divergence. This requires the objective function to obey 1-Lipschitz constraint, which is originally enforced by weight clipping. Guljarani et al. have shown a more efficient constraint in a form of gradient penalty, resulting in WGAN-GP model [11].

It should be noted the development of generative models is driven mostly by advances in visual data generation. It is due to the relative ease of qualitative evaluation of such data – either by humans or by using special metrics, such as Inception score [12]. A recent approach, WaveGAN, has shown that it is possible to produce realistic audio data with GAN [13]. In this case, the data is evaluated both by humans (and a cat) as well as the Inception score obtained by training the audio classification model operating on spectrograms of audio samples. Additionally, WaveGAN utilizes a specific regularization technique for discriminator, called phase shuffle, which performs random phase perturbation of activations of 1D convolutional layers.

### IV. GENERATIVE MODEL

#### A. GAN architecture

Proposed GAN architecture for generation of ECG signals (which we call ECG-GAN) is inspired by WaveGAN model by Donahue et al. [13]. We train the generative model using Wasserstein GAN with Gradient Penalty variant (WGAN-GP). As real input, we use ECG signals from PTB-XL database [14]. The database contains over 21k samples 10s length each, sampled at 500 Hz, available for each of 12 diagnostic leads, along with rich metadata. For generative training we use signals from aVL lead only, due to its diagnostic significance as well as to reduce dataset complexity. Signals are scaled independently to $<-1,1>$ range. Each signal consists of 5k data points. The generator network input is 1D random vector of $z$ length with each element sampled from normal distribution. This latent vector is mapped onto dense linear layer followed by several layers of transposed convolution with leaky ReLU activation and exponentially decaying number of filters. After the last convolution, we crop the output symmetrically at both ends to 5k width and process it by the hyperbolic tangent function. Apart from the necessity of having output of specific length, cropping allows the network to generate signal before and after effective range, which minimizes signal distortions caused by padding. In all but last convolution we use batch normalization as a regularization mechanism. The critic network is made of several convolution block with ReLU activation and exponentially increasing number of filters. The last layer maps the signal into a single linear output, to produce unbounded critic score. Since batch normalization is unadvised for a critic network due to batch correlation phenomenon [11], we use the phase shuffle, as suggested by Donahue et al. [13], instead. We train networks to minimize Wasserstein distance using batches of size 64. For both networks we use Adam optimizer, with 5 critic updates per each generator update.

#### B. Evaluation of generative model

Qualitative evaluation of samples produced by generative models is preferably done by human judges, since quantitative metrics are known to be largely uncorrelated to subjective evaluation [15]. Qualitative analysis is associated, as one may expect, with many difficulties, both logistic and financial. There is, however, a widely used metric known as Inception score, that uses pretrained Inception neural network model [12] to provide an evaluation highly correlated to human evaluation. Given input dataset $x$, and model output $y$, it is defined as: $\exp(\mathbb{E}_x \mathbf{KL}(p(y|x) \parallel p(y)))$, where **KL** is Kullback-Leibler divergence. Samples that represent meaningful objects should have conditional label distribution $p(y|x)$ similar to marginal distribution $p(y)$. Despite known shortcomings [16], Inception score remains a popular choice of evaluation metric for visual generative models. An analogous model was then proposed for adversarial audio synthesis [13], using classifier trained on spectrograms of audio samples. We propose a similar model, trained on spectrograms of ECG signals. We also argue that Inception score is more suitable here than in, for example, image analysis, since human brain (especially visual and auditory cortices) has evolved to analyze "natural" data representing real objects rather than "unnatural" ones such as ECG. Therefore, it is more difficult to generate obviously fake

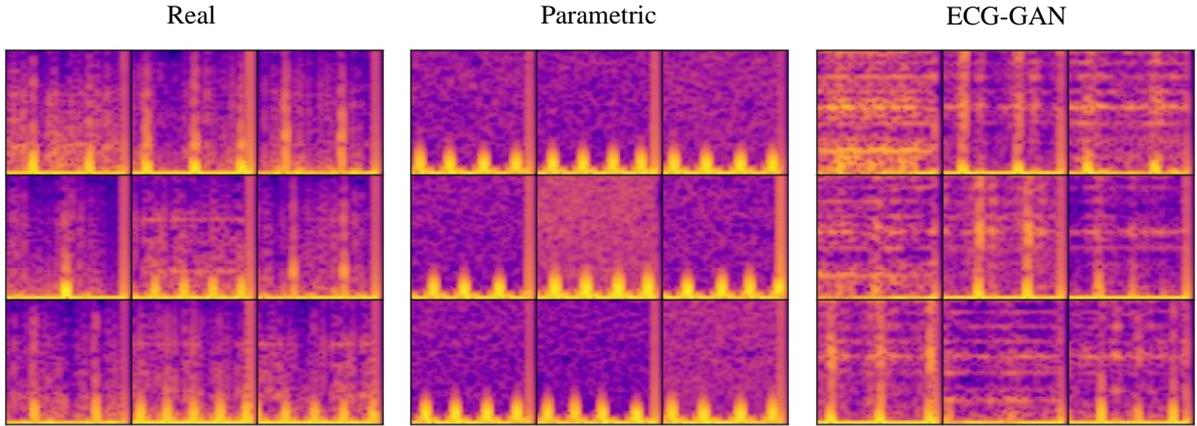

Figure 2. Random samples of spectrograms of various ECG datasets. Horizontal axes represent time, while vertical axes represent frequency (on the Mel scale). Each spectrogram represents a full-length signal (5000 data points).

examples that would obtain high Inception score anyway [16].

Our "Inception-like" classifier is a 2D convolutional network operating on 64x64 spectrograms. Each spectrogram is produced by performing short-time Fourier Transform on 1024 data points and mapping resulting spectrogram onto Mel scale. While Mel scale is mostly known for its application for audio data, it is also used for ECG signal preprocessing for neural networks [5], [17]. Resulting spectra are normalized to zero mean and unit variance, clipped to 3 standard deviations and rescaled to $<-1,1>$ range. Each spectrogram has 0 to 5 output labels assigned, indicating diagnostic classes from PTB-XL. We end the classifier network with sigmoid activation and train it using binary crossentropy loss, effectively turning single multilabel classification problem into a set of binary classification tasks. The network is trained for 100 epochs with Adam optimizer and batch size 64. We validate the model after each epoch and save the one with the lowest validation loss. The final model achieves mean average precision 0.63 and mean average recall 0.41 for validation set. To obtain Inception score, we first get the output of the network for dataset $x$ and calculate $p(y|x)$ by summing up probabilities $p(y_i|x)$ for each label and normalizing the sum to 1. Then we calculate the score using the Inception equation.

In addition to the Inception score, we use two additional metrics to verify whether the model result is not degenerated to one of pathological cases. The first metric, $|D|_{\text{self}}$, measures average Euclidean distance between the signal and its nearest neighbor. Low value would indicate that the model generates very similar samples, experiencing mode collapse. Another metrics $|D|_{\text{train}}$, measures average distance between signals and their closest neighbors in the training set. Low value is a symptom of model overfitting – the model learned to reproduce samples from the training set, unable to produce novel ones.

## V. DENOISING MODEL

### A. Denoising model architecture

Proposed denoising model is a deep denoising autoencoder (DAE) [18] operating in time-domain. It consists of two blocks – encoder and decoder – connected together. The encoding block consists of 4 convolution layers with increasing number of filters. It is followed by the decoding block of 4 transposed convolution layers. The output is cropped to match the input signal length. Each layer has leaky ReLU activation. The baseline model uses no regularization. The detailed description of the model can be found in Table 6. Architecture of the encoder block is almost the same as the critic model (apart from the phase shuffle and the last convolutional and dense layers), allowing us to experiment with transfer learning by using pretrained critic layers as the encoder.

We train the denoising neural network in time-domain to reconstruct initial signal, given signal with added artificial noise with strength $\gamma = 1$. The baseline model is trained using a mix of real (17k samples) and synthetic (100k) signals. Models are evaluated independently on real (2k) and synthetic (100k) test datasets. We use Adam optimizer with MSE loss and hyperparameter values the same as during generative model training. Models are validated after each epoch and the one with the best validation loss is used for evaluation.

### B. Noise model

Typically, denoising autoencoders use simple noise models, like salt-an-pepper noise, additive white gaussian noise or black (masking) noise [19]. They are applied to the input signal while the autoencoder learns to reconstruct the original signal by finding a certain hidden representation. This hidden representation should encapsulate useful structural features of the data. However, we argue that in case of ECG signal, the specificity and complexity of the noise should be also considered during signal reconstruction. Therefore, we propose to use a more sophisticated noise model, based on distortions present in real ECG. There are multiple known sources of noise in real ECG signal [1], each having a specific frequency and power range. Our noise model is a parametric one that takes into account three of the most common noise sources, namely, baseline wander ($a_w$), motion artifacts ($a_m$) and power line interference ($a_p$), multiplied by strength parameter $\gamma$:

$$x_{noised}(t) = x(t) + \gamma(a_w(t) + a_m(t) + a_p(t))$$

Baseline wander [20] is a low-frequency component, with usual frequency up to 0.6 Hz, caused mainly by respiration.

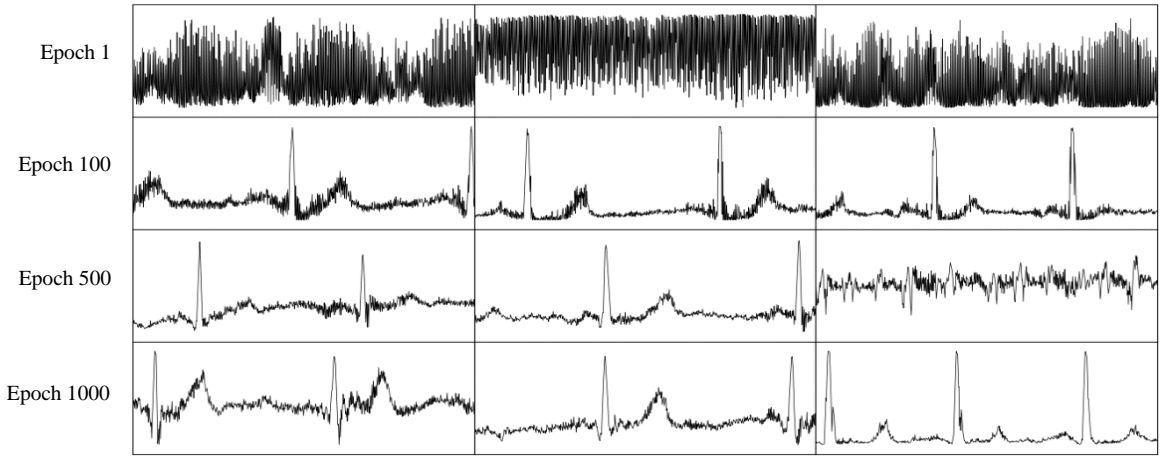

Figure 3. Samples of ECG signal generated by ECG-GAN during various stages of training. Each column shows samples generated using the same latent vector *z*. Visible are first 1024 data points of each signal. Vertical axes are limited to <-1,1> range.

We model it as a sine function with random frequency sampled uniformly from 0-0.5 Hz range and amplitude sampled randomly from 0-0.3 range (assuming that the initial signal has -1 to 1 amplitude). The motion artifact component is caused by variation of electrode capacitor due to body motion. It covers a wide range of frequencies (0.5 to 120 Hz) overlapping with the usable spectrum of ECG signal. We model it by the modulated chirp signal, i.e. sinusoidal waveform with increasing frequency and amplitude modulated by another sine wave [21]. The power line interference is produced by differences in electrode impedances as well as stray currents through the patient and the cables. It mostly has a frequency equal to that of electrical system in given country, therefore we model it as a sine waveform with 50 Hz, which is the standard frequency in most countries [22]. The amplitude of this component is 0-0.1 of the source signal amplitude.

### C. Evaluation of denoising models

When evaluating filtering methods, one should take into consideration a purpose of filtering. Therefore, we use three unrelated metrics to evaluate denoising models. The first one is mean squared error (MSE) which is the most general method and is also used as a loss function during the training of neural networks. The second metric is signal-to-noise ratio, which allows to measure a relative strength of the recovered signal to the noise. The third metric reflects the quality of the ECG signal from the diagnostic point of view. One of the most essential diagnostic activities for time-domain ECG analysis is identification of QRS complexes, which serve as a basis to determine heart rate variability. We therefore introduce $\delta_{HR}$ metric, defined as the mean absolute error of the heart rate (HR) value between noise-free and denoised signal. The heart rate is calculated as the reciprocal of average length of the cardiac cycle, which, in turn, is calculated as the distance (in seconds) between neighboring QRS peaks. For QRS detection, we use an open-source QRS detector developed by Sznajder and Łukowska [23]. It is based on Pan-Tompkins algorithm [24] which is one of the most common algorithms used for this purpose. Signal amplitudes were scaled to the detector operational range.

### VI. EXPERIMENTAL SETUP

In the first phase of the experiment we train and evaluate ECG-GAN model by calculating Inception score as well as $|D|_{\text{self}}$ and $|D|_{\text{train}}$ metrics. We compare the proposed model with the reference parametric model of ECG signal introduced by McSharry et al. [25]. It is a dynamical model that incorporates various morphological parameters (such as heart rate or PQRST points locations) as well as the measurement process details (like sampling frequency or measurement noise). According to authors, the quality of the model allows it to be used as a benchmark for biomedical signal processing techniques. In addition to the proposed and reference models, we calculate metrics for training and test dataset as well to obtain a reference point. The total training length of ECG-GAN is 1k epochs, which takes around 3 days on a machine with Intel Xeon E5-2640 CPU and Nvidia Tesla K20m GPU.

During the second phase, we train and evaluate denoising models. We use two reference methods, namely the bandpass filter and the wavelet filter. The bandpass filter consists of high- and low- pass filters that cut out frequencies outside of 0.05-30 Hz range. The wavelet filter uses Undecimated Wavelet Transform to decompose the signal. As a mother wavelet we use D6 wavelet from Daubechies family. Its shape is similar to QRS component, making it a viable choice for ECG signal filtering. We also perform a series of experiments by modifying the baseline denoising model. We train models with various amounts of synthetic data as well as combined synthetic + real data. This allows us to verify whether using the data from the generative model has positive effect on the denoising model. We also investigate whether the impact of synthetic data can be substituted with regularization, by training networks with phase shuffle added before each convolution in encoding block. Finally, we use the pretrained critic network as an encoder block of the denoiser to check whether the model will be able to transfer features learned during GAN training to the new task.

Table 1. Quantitative results of ECG datasets analysis.

| Dataset | Inception score | $|D|_{\text{self}}$ | $|D|_{\text{train}}$ |
| --- | --- | --- | --- |
| Real (train) | $1.319 \pm 0.002$ | 18.11 | 0.0 |
| Real (test) | $1.314 \pm 0.004$ | 18.01 | 18.09 |
| Parametric | $1.090 \pm 0.037$ | 18.64 | 22.75 |
| ECG-GAN | $1.268 \pm 0.004$ | 17.34 | 20.49 |

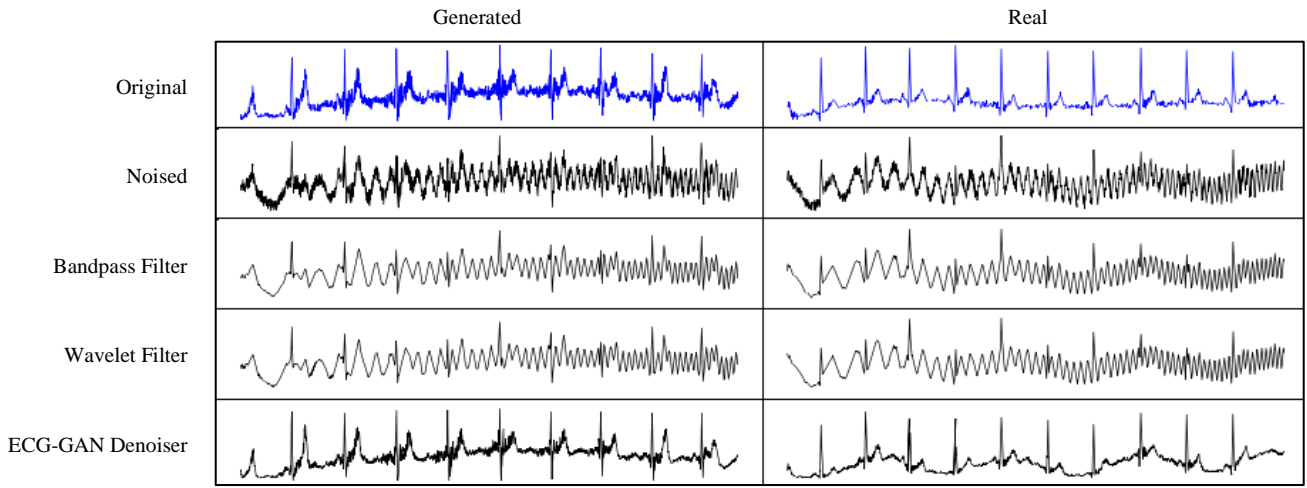

Figure 4. Comparison of filtering methods. Left column contains results for synthetic signal generated with ECG-GAN, while right column – a real one.

## VII. RESULTS

### A. Quality of generative model

Visual examples of signals generated by ECG-GAN are presented in Appendix C. Quantitative results of datasets comparison are presented in Table 2. Signals generated by ECG-GAN obtained much higher score than parametric model, while being largely unrelated to the training set, as indicated by high $|D|_{train}$ value. An informal human analysis was performed, which confirms these results. Synthetic signals had a wider frequency range compared to parametric data, which lack spectral variability, as seen in spectrograms in Figure 2. One can also easily identify characteristic elements of ECG on generated data, such as QRS complex as well as P and T waves.

### B. Quality of denoising models

The collected results are shown in Table 2. Proposed denoising model (ECG-GAN Denoiser) outperformed reference methods on both generated and real test sets in terms of each metric. Interestingly, networks trained exclusively on real or synthetic data (rows 6 and 8, respectively) performed worse than the network trained on mixed data. A more detailed analysis of impact of synthetic training dataset on test set performance can be seen in Figure 5. Mixing real dataset with generated data helps to minimize drawbacks of both. Regularization in a form of phase shuffle improves performance of real data -trained network. Nevertheless, the effect of generated data is more profound than simple regularization. By combining the phase shuffle with the baseline model, we achieve the best performance for real signals. Nevertheless, we do not observe a significant beneficial effect of reusing the pretrained critic network in denoising model.

Table 2. ECG denoising results for various methods. In case of $S/N$, the higher value is better. For MSE and $\delta_{HR}$ – the lower, the better. The (none) row indicates metrics for the noisy signal before any filtering.

| Denoising method | Performance (generated test set) | | | Performance (real test set) | | |
|---|---|---|---|---|---|---|
| | MSE | $S/N$ | $\delta_{HR}$[Hz] | MSE | $S/N$ | $\delta_{HR}$[Hz] |
| (none) | 0.139 | 3.243 | 0.911 | 0.127 | 2.332 | 0.912 |
| Bandpass filter | 0.141 | 3.185 | 0.846 | 0.122 | 2.532 | 0.789 |
| Wavelet filter | 0.141 | 3.183 | 0.850 | 0.123 | 2.516 | 0.805 |
| ECG-GAN Denoiser | 0.018 | 11.459 | **0.444** | 0.028 | 8.185 | 0.355 |
| ECG-GAN Denoiser (phase shuffle) | **0.017** | **11.735** | 0.450 | **0.026** | **8.395** | **0.349** |
| ECG-GAN Denoiser (real data only) | 0.041 | 7.963 | 0.541 | 0.031 | 7.730 | 0.448 |
| ECG-GAN Denoiser (real data only + phase shuffle) | 0.038 | 8.293 | 0.609 | 0.027 | 8.260 | 0.535 |
| ECG-GAN Denoiser (generated data only) | 0.028 | 9.559 | 0.570 | 0.040 | 6.778 | 0.554 |
| ECG-GAN Denoiser (pretrained) | 0.038 | 8.587 | 0.727 | 0.047 | 6.100 | 0.627 |

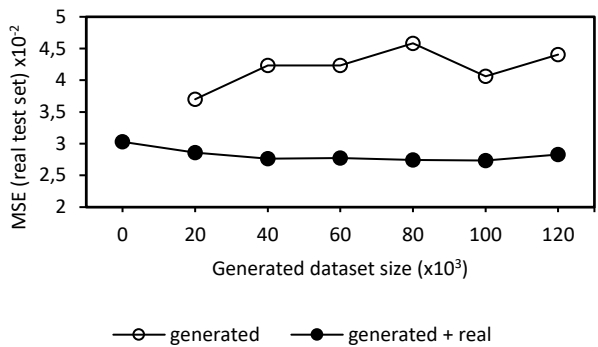

Figure 5. Denoising model performance for various training set sizes.

## VIII. CONCLUSIONS

We have presented state-of-the-art deep model for ECG signal denoising, trained with the help of GAN. We have thereby demonstrated that generative adversarial models can serve as a powerful data augmentation technique that improves both the quality and the quantity of the dataset. Moreover, the effect of generated data goes beyond the simple regularization, which is the case for simple data transformations [26].

Generative model presented in this paper, despite producing "good enough" training data, is quite simple and lacks more advanced features that can be found in modern GANs for image generation. One of the most promising directions is a "style transfer" technique that provides control over stochastic variation of generated data [27]. This would allow, for example, to generate a signal from the specific lead. We notice that our Inception network returns low scores compared to image classifiers with similar number of output labels. This is due to the fact that only a single lead (aVL) is used for the classifier training. Models that utilize data from all 12 leads available in PTB-XL database are known to achieve much higher precision and recall [28]. A more accurate classifier would return a more credible Inception score. Similarly, it is reasonable to expect that the performance of the denoising model would benefit from data from multiple leads.

## ACKNOWLEDGEMENT

Author would like to thank Systems Engineering Advanced Studies Centre (SEC) team at Cybernetics Faculty, Military University of Technology, for providing computing infrastructure and excellent support.

# A    NETWORKS ARCHITECTURE

Table 3. ECG-GAN Generator architecture.

| Layer | Kernel Size | Output Shape |
|---|---|---|
| Input $z \sim \text{Uniform}(-1,1)$ | | $(n, 100)$ |
| Dense | $(100, 128d)$ | $(n, 128d)$ |
| Reshape | | $(n, 8, 16d)$ |
| Trans Conv1D (stride=4) | $(25, 16d, 8d)$ | $(n, 32, 8d)$ |
| Batch Norm | | $(n, 32, 8d)$ |
| LReLU ($\alpha = 0.2$) | | $(n, 32, 8d)$ |
| Trans Conv1D (stride=4) | $(25, 8d, 4d)$ | $(n, 128, 4d)$ |
| Batch Norm | | $(n, 128, 4d)$ |
| LReLU ($\alpha = 0.2$) | | $(n, 128, 4d)$ |
| Trans Conv1D (stride=4) | $(25, 4d, 2d)$ | $(n, 512, 2d)$ |
| Batch Norm | | $(n, 512, 2d)$ |
| LReLU ($\alpha = 0.2$) | | $(n, 512, 2d)$ |
| Trans Conv1D (stride=4) | $(25, 2d, d)$ | $(n, 2048, d)$ |
| Batch Norm | | $(n, 2048, d)$ |
| LReLU ($\alpha = 0.2$) | | $(n, 2048, d)$ |
| Trans Conv1D (stride=4) | $(25, d, 1)$ | $(n, 8192, 1)$ |
| Cropping | | $(n, 5000, 1)$ |
| Tanh | | $(n, 5000, 1)$ |

Table 5. ECG-GAN Critic architecture.

| Layer | Kernel Size | Output Shape |
|---|---|---|
| Input | | $(n, 5000, 1)$ |
| Conv1D (stride=4) | $(25, 1, 1)$ | $(n, 1250, 1)$ |
| Phase Shuffle | | $(n, 1250, 1)$ |
| LReLU ($\alpha = 0.2$) | | $(n, 1250, 1)$ |
| Conv1D (stride=4) | $(25, 1, d)$ | $(n, 313, d)$ |
| Phase Shuffle | | $(n, 313, d)$ |
| LReLU ($\alpha = 0.2$) | | $(n, 313, d)$ |
| Conv1D (stride=4) | $(25, d, 2d)$ | $(n, 79, 2d)$ |
| Phase Shuffle | | $(n, 79, 2d)$ |
| LReLU ($\alpha = 0.2$) | | $(n, 79, 2d)$ |
| Conv1D (stride=4) | $(25, 2d, 4d)$ | $(n, 20, 4d)$ |
| Phase Shuffle | | $(n, 20, 4d)$ |
| LReLU ($\alpha = 0.2$) | | $(n, 20, 4d)$ |
| Conv1D (stride=4) | $(25, 4d, 8d)$ | $(n, 5, 8d)$ |
| Phase Shuffle | | $(n, 5, 8d)$ |
| LReLU ($\alpha = 0.2$) | | $(n, 5, 8d)$ |
| Reshape | | $(n, 40d)$ |
| Dense | $(40d, 1)$ | $(n, 1)$ |

Table 4. ECG-GAN Inception architecture.

| Layer | Kernel Size | Output Shape |
|---|---|---|
| Input | | $(n, 64, 64, 1)$ |
| Conv2D (stride=2) | $(3, 3, 1, 64)$ | $(n, 32, 32, 64)$ |
| Batch Norm | | $(n, 32, 32, 64)$ |
| ReLU | | $(n, 32, 32, 64)$ |
| MaxPool2D (stride=2) | | $(n, 16, 16, 64)$ |
| Conv2D (stride=2) | $(3, 3, 64, 64)$ | $(n, 8, 8, 64)$ |
| Batch Norm | | $(n, 8, 8, 64)$ |
| ReLU | | $(n, 8, 8, 64)$ |
| MaxPool2D (stride=2) | | $(n, 4, 4, 64)$ |
| Conv2D (stride=2) | $(3, 3, 64, 64)$ | $(n, 2, 2, 64)$ |
| Batch Norm | | $(n, 2, 2, 64)$ |
| ReLU | | $(n, 2, 2, 64)$ |
| MaxPool2D (stride=2) | | $(n, 1, 1, 64)$ |
| Reshape | | $(n, 64)$ |
| Dense | $(64, 5)$ | $(n, 5)$ |
| Sigmoid | | $(n, 5)$ |

Table 6. ECG-GAN Denoiser architecture.

| Layer | Kernel Size | Output Shape |
|---|---|---|
| Input | | $(n, 5000, 1)$ |
| Conv1D (stride=4) | $(25, 1, 1)$ | $(n, 1250, 1)$ |
| LReLU ($\alpha = 0.2$) | | $(n, 1250, 1)$ |
| Conv1D (stride=4) | $(25, 1, d)$ | $(n, 313, d)$ |
| LReLU ($\alpha = 0.2$) | | $(n, 313, d)$ |
| Conv1D (stride=4) | $(25, d, 2d)$ | $(n, 79, 2d)$ |
| LReLU ($\alpha = 0.2$) | | $(n, 79, 2d)$ |
| Conv1D (stride=4) | $(25, 2d, 4d)$ | $(n, 20, 4d)$ |
| LReLU ($\alpha = 0.2$) | | $(n, 20, 4d)$ |
| Trans Conv1D (stride=4) | $(25, 4d, 4d)$ | $(n, 80, 4d)$ |
| LReLU ($\alpha = 0.2$) | | $(n, 80, 4d)$ |
| Trans Conv1D (stride=4) | $(25, 4d, 2d)$ | $(n, 320, 2d)$ |
| LReLU ($\alpha = 0.2$) | | $(n, 320, 2d)$ |
| Trans Conv1D (stride=4) | $(25, 2d, d)$ | $(n, 1280, d)$ |
| LReLU ($\alpha = 0.2$) | | $(n, 1280, d)$ |
| Trans Conv1D (stride=4) | $(25, d, 1)$ | $(n, 5120, 1)$ |
| LReLU ($\alpha = 0.2$) | | $(n, 5120, 1)$ |
| Cropping | | $(n, 5000, 1)$ |
| Tanh | | $(n, 5000, 1)$ |

## B  TRAINING HYPERPARAMETERS

Table 7. Training hyperparameters.

| Hyperparameter | Value |
| --- | --- |
| Batch size ($n$) | 64 |
| Model dimensionality ($d$) | 16 |
| Gradient penalty ($\lambda$) | 10 |
| $C$ updates per $G$ update | 5 |
| Phase shuffle | 2 |
| Adam learning rate ($\alpha$) | 1e-4 |
| Adam 1st moment decay ($\beta_1$) | 0.9 |
| Adam 2nd moment decay ($\beta_2$) | 0.999 |

## C  EXAMPLES OF GENERATED SIGNALS

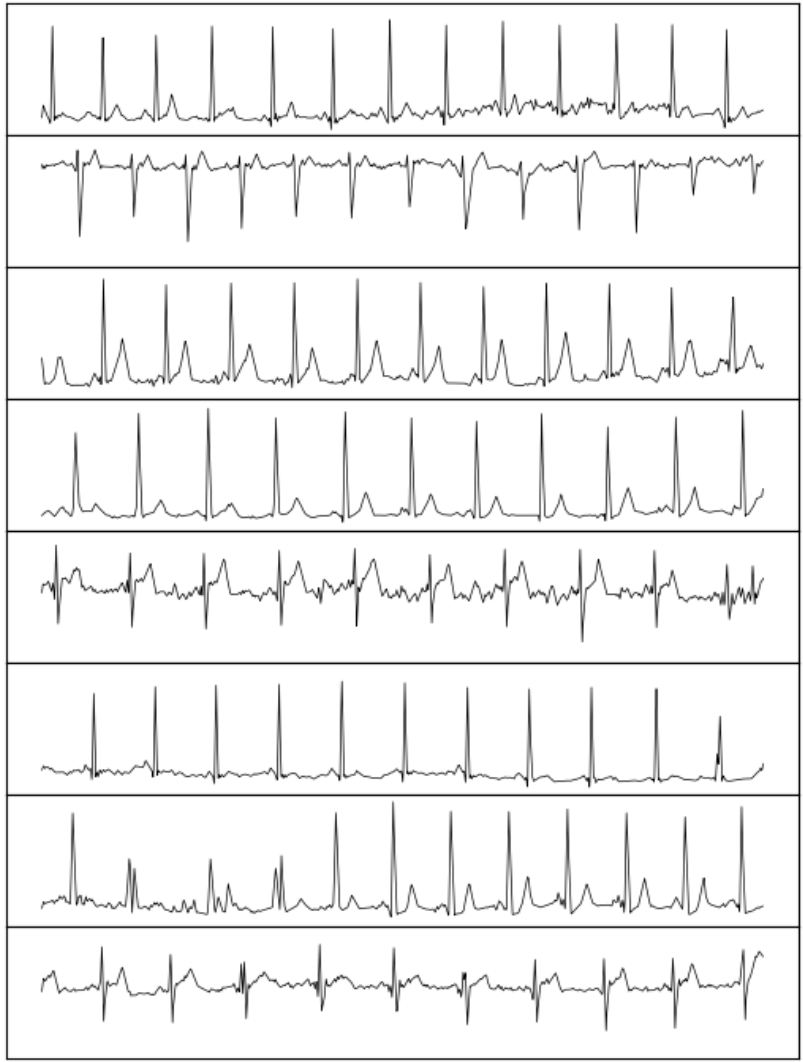

Figure 6. Examples of signals generated by ECG-GAN. Each sample represents 10 seconds of simulated heart activity. For better visibility, signals are processed with lowpass filter to cut out >30Hz noise.